\documentclass{article}

\usepackage{arxiv}

\usepackage[utf8]{inputenc} 
\usepackage[T1]{fontenc}    
\usepackage{hyperref}       
\usepackage{url}            
\usepackage{booktabs}       
\usepackage{amsfonts}       
\usepackage{nicefrac}       
\usepackage{microtype}      
\usepackage{lipsum}		
\usepackage{graphicx}
\usepackage{natbib}
\usepackage{doi}

\title{T4PdM: a Deep Neural Network based on the Transformer Architecture for Fault Diagnosis of Rotating Machinery}

\author{ \href{https://orcid.org/0000-0000-0000-0000}{\includegraphics[scale=0.06]{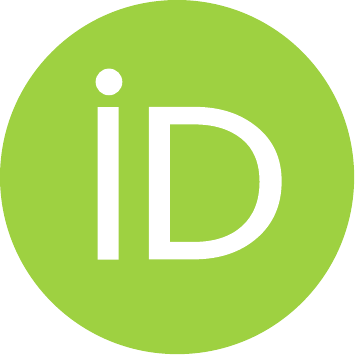}\hspace{1mm}Erick Giovani Sperandio Nascimento} \\
	SENAI CIMATEC\\
	Salvador, Bahia, 41650-010, Brazil  \\
	\texttt{erick.sperandio@fieb.org.b} \\
	\And
	\href{https://orcid.org/0000-0000-0000-0000}{\includegraphics[scale=0.06]{orcid.pdf}\hspace{1mm}Julian Santana Liang} \\
	SENAI CIMATEC\\
	Salvador, Bahia, 41650-010, Brazil \\
	\texttt{julian.liang@fbter.org.br} \\
	\And
	\href{https://orcid.org/0000-0000-0000-0000}{\includegraphics[scale=0.06]{orcid.pdf}\hspace{1mm}Ilan Sousa Figueiredo} \\
	SENAI CIMATEC\\
	Salvador, Bahia, 41650-010, Brazil \\
	\texttt{ilan.figueiredo@fieb.org.br} \\
	\And
	\href{https://orcid.org/0000-0000-0000-0000}{\includegraphics[scale=0.06]{orcid.pdf}\hspace{1mm} Lílian Lefol Nani Guarieiro} \\
	SENAI CIMATEC\\
	Salvador, Bahia, 41650-010, Brazil  \\
	\texttt{lilian.guarieiro@fieb.org.br} \\
}

\renewcommand{\shorttitle}{\textit{arXiv} Template}

\hypersetup{
pdftitle={A template for the arxiv style},
pdfsubject={q-bio.NC, q-bio.QM},
pdfauthor={Erick Giovani Sperandio Nascimento, Julian Santana Liang, Ilan Sousa Figueredo, Lílian Lefol Nani Guarieiro},
pdfkeywords={First keyword, Second keyword, More},
}

\begin{document}
\maketitle

\begin{abstract}
	Deep learning and big data algorithms have become widely used in industrial applications to optimize several tasks in many complex systems. Particularly, deep learning model for diagnosing and prognosing machinery health has leveraged predictive maintenance (PdM) to be more accurate and reliable in decision making, in this way avoiding unnecessary interventions, machinery accidents, and environment catastrophes. Recently, Transformer Neural Networks have gained notoriety and have been increasingly the favorite choice for Natural Language Processing (NLP) tasks. Thus, given their recent major achievements in NLP, this paper proposes the development of an automatic fault classifier model for predictive maintenance based on a modified version of the Transformer architecture, namely T4PdM, to identify multiple types of faults in rotating machinery. Experimental results are developed and presented for the MaFaulDa and CWRU databases. T4PdM was able to achieve an overall accuracy of 99.98\% and 98\% for both datasets, respectively. In addition, the performance of the proposed model is compared to other previously published works. It has demonstrated the superiority of the model in detecting and classifying faults in rotating industrial machinery. Therefore, the proposed Transformer-based model can improve the performance of machinery fault analysis and diagnostic processes and leverage companies to a new era of the Industry 4.0. In addition, this methodology can be adapted to any other task of time series classification.
\end{abstract}


\section{Introduction}
Vibrations are oscillating movements of devices around their balanced position. Any change in signal amplitude or frequency may indicate that the performance or quality of the device has been compromised. The wearing caused by mechanical friction alters the characteristics of the vibration signals, which allows the detection of an anomaly. A vibration analysis of rotating machines is popular because it enables the execution of tests to detect errors without having to stop the production line, i.e., while the system is in operation \citep{Vishwakarma2017}. 

As a result of the rapidly reducing sensor and data storage costs, as well as the increasing computer capacity, smarter systems can be developed for early detection of machine failures \citep{Lu2017}. As a consequence, it is feasible to expand the structures that assist in the maintenance of these equipment using intelligent techniques \citep{Lee2019}. Much more than a trend, smart manufacturing is a vital issue for industries that want to be productive and competitive in the face of the fourth industrial revolution. New technologies allow the continuous improvement of machine conditions and control of operation. Through the developments of Artificial Intelligence (AI), models based on Machine Learning (ML), and more recently Deep Learning (DL) technologies, modern fault diagnosis techniques have emerged as a powerful tool for predictive maintenance (PdM). The main advantage of ML approaches is the ability to handle high-dimensional data (e.g.: multivariate time series) and to extract hidden relationships within data in complex and dynamic environments. In this reality, the application of AI aims to seek an optimized production process with higher quality and reduced time and costs, thus generating a chance to use assembly line data and obtain feedback in a timely manner and driving the improvement of production processes to be continued with fewer possible interventions \citep{Wuest2016}.

According to the current research scenario, fault diagnosis methods based on Convolutional Neural Networks (CNN) and Recurrent Neural Networks (RNN) have been well studied in many fields of knowledge. On the other hand, despite the fact that models based on the Transformer architecture have demonstrated a paramount success in the field of Natural Language Processing (NLP), they have not been widely studied, applied, and evaluated for time series classification problems, especially in the field of machinery fault diagnosis. The Transformer architecture allows more parallelization, avoids duplication processing, and relies on the attention mechanism \citep{Vaswani2017}. 

In this view, this research proposes a Transformer-based fault diagnosis model, namely T4PdM, in order to classify failures in rotating machines. Therefore, the Transformer architecture is applied to build a deep neural network (DNN) to identify and classify the type of failures in a rotating machine for different rotational speed, load level, and severities with the goal of evaluating and assessing the performance of the presented model using two publicly available datasets for machinery fault diagnosis. Thus, T4PdM was developed, applied, and evaluated in relation to previous studies. In order to compare the performance of the proposed model, the quantitative classification metrics of accuracy, precision, recall, and f1-score were calculated for two real-world and publicly available databases: Machinery Fault Database (MaFaulDa \footnote{\url{http://www02.smt.ufrj.br/~offshore/mfs/page_01.html}}) and Case Western Reserve University (CWRU\footnote{\url{https://engineering.case.edu/bearingdatacenter/download-data-file}}) [6]). 

The rest of this work is structured as follows: Section 2 describes the literature review. Section 3 presents the methodology used to build, develop, and evaluate the proposed model. Results and discussions are provided in Section 4, and finally in Section 5, conclusions and future works are presented.

\section{Review of literature}
\label{sec:headings}

Because of the fast advancement of machinery monitoring technologies, the industry's thinking, planning, and development have undergone significant changes. At present, most production systems have adopted new technologies for digital transformation to increase output through enhanced maintenance. However, because of the challenges and high costs associated with the digital transformation processes, several areas such as rotating machinery failure classification are having difficulty to adopt it in the production line. Among the failures affecting production machines, structural rotor fault is a prevalent and straightforward fault, which includes unbalance, misalignment, looseness, etc. Notwithstanding, AI provides a promising solution to classify the imminence of these failures. As an example, \citet{Baptista2018} has developed AI approaches and compared them to statistical methods in the prediction of when devices would fail. The results have revealed that AI approaches are better than statistical methods. 

The literature reports that conventional feature extraction in the time domain fails to identify symptom parameters because of anomalies in the raw data \citep{Xue2013}. Therefore, the use of frequency domain is more widely adopted for feature extraction than time domain in conjunction with Deep Learning models. Because of data scarcity and imbalance issues, \citet{Nath2021} has generated a subsampled dataset incorporating Distinctive Frequency Components (DFC) addressed by an incremental method using soft-Dynamic Time Warping (soft-DTW), which enhanced the AI model by the weighing scheme based on Fault Information Content (FIC). 

CNN models have been successfully applied to classify machinery faults because they can automatically extract the main characteristics of raw vibration data. The work of \citet{Kolar2020} has presented a Multi-Channel Deep CNN (MC-DCNN) configuration for the state classification of rotating machinery. The authors have proposed a system that automatically extracts signal characteristics by importing the original data from a three-axis accelerometer as a high-definition one-dimensional image, which may eliminate expert knowledge in vibration signal preprocessing. Multivariate raw signal data is divided into univariate in a way that each channel presents input in a feature learning stage. Recently, \citet{Souza2021} has presented a model called PdM-CNN, which uses a specialized combination of fast Fourier transform and 1D-CNN for the automatic classification of faults in unbalanced data from rotating machine using data from only one vibration sensor. The authors have applied the model to the MaFaulDa and CWRU public datasets and demonstrated superior performance than previous works in the task of fault classification in different operating scenarios of rotational speed, load levels, and severities.

Recently, RNN has been applied to fault diagnosis and achieved superior performance than other equivalent CNN-based approaches, because RNN architectures have advantages in harnessing temporal information from time series vibration signals because of the recurrent hidden layers \citep{Jiang2018}. In view of this, the work of \citet{Zhang2021} has constructed an image from one-dimensional vibration signal, used a linear layer to increase the dimensionality of each signal segment, and then developed an extraction method based on Gated Recurrent Unit (GRU), learning and capturing representative features, and tested it with two databases: CWRU and Self-Priming Centrifugal Pump (SPCP). The work of \citet{Nath2021} has proposed an early classification model based on Long Short-Term Memory (LSTM) and GRU architectures for structural rotor fault diagnosis. The model achieved an accuracy of 99.5 with the Meggitt dataset and an accuracy of 98.32 with the MaFaulDa dataset.
More recently, because of its success in the area of NLP, Transformer architecture has been also applied to other areas, such as time series classification and, more specifically, fault diagnosis for PdM tasks. Although there are few works, it is possible to perceive an increasing interest in this area of research. For example, the work of \citet{Jin2021} has introduced a token sequence generation method to develop a Transformer architecture model to recognize fault modes of bearings using raw vibration signals of one-dimensional (1D) data format. The distribution form of the feature vectors was compared with the traditional CNN and RNN models, and the Transformer models revealed the best intra-class compactness and inter-class separability. On the other hand, \citet{Bao2021} has applied a Transformer-based model for fault diagnosis using the short-time Fourier transform to convert the one-dimensional fault signal into a two-dimensional image. Fault classification may achieve an accuracy of 98.45\%, according to experimental results. Another work (\citep{Ribeiro}) has proposed a Time-Frequency Transformer (TFT) model, based on a tokenizer and encoder module designed to extract effective abstractions from the time-frequency representation (TFR) for fault diagnosis, where they were obtained by applying the synchronized wavelet transform (SWT) to the raw vibration data. The authors have used datasets that were collected from Accelerated Bearing Life Tester (ABLT-1A).

As we can notice, few works aim to evaluate, test, and improve Transformer architecture for fault diagnosis of rotating machinery. Thus, inspired by \citet{Marins2017}, our work aims to innovate by developing a modified version of the Transformer architecture, called T4PdM, for fault classification and diagnosis of industrial rotating machines also using vibration data from only one vibration sensor, which is a more commonly adopted configuration in the industry \citep{Marins2017}, in this way evaluating its performance in two different and public datasets. 

\section{Methodology}
\label{sec:headings}
This section describes the methodology employed to develop the proposed T4PdM model: (i) dataset and feature selection (FS), (ii) feature extraction, (iii) the proposed T4PdM model for fault diagnosis in rotating machines, and (iv) evaluation performance of the proposed model and comparison with other DL models developed and applied for the same task. 

\subsection{Dataset and pre-processing}
Two real-world open datasets were used in this paper: (I) MaFaulDa  and (ii) CWRU. These datasets are extensive enough and include multiple faults from the rotor and bearing systems. Each fault has different severity levels and different speeds of operation.

 The MaFaulDa dataset is composed of multivariate time series with five fault conditions - (i) imbalanced operation, (ii) horizontal misalignment, (iii) vertical misalignment, and (iv) underhang and (v) overhang bearing fault (outer race fault, ball fault, and cage fault) - with three imbalanced load values of 0g, 6g, 20g, and 35g \citep{Ribeiro}. Only data collected in three unidirectional accelerometers (similar to one three-axis accelerometer) were selected in this paper, in the same way as \citet{Souza2021} has done, since this setting is one of the most common layout adopted for monitoring machinery in the industry. Thus, the overhang bearing data was discarded.
 The CWRU dataset has three fault types - ball fault, outer race fault, and inner race fault - and one normal condition. The dataset was divided into four categories: 48 KHz baseline, 48 KHz drive end fault, 12 KHz drive end fault, and 12KHz fan end fault \citep{Marins2017}. We selected from the CWRU dataset only the aggregated data from the 48K drive end fault, which corresponds to the 48 KHz sample rate, and the one unidirectional accelerometer mounted at the drive end bearing position.

\subsection{Feature extraction}
The phase of feature extraction is composed of two steps. The first one is to convert the data from time domain to frequency domain through the Fast Fourier Transform (FFT). FFT uses regular intervals by subsampling data points from a continuous stream of data. Thus, the FFT spectrum was applied in each segment of the data bringing the data from the time domain to the frequency domain. The second step is to reduce the dimensionality of the dataset through Principal Component Analysis (PCA) to low-dimensional and non-redundant data. In this way, 4,500 components were extracted out of 7,500 original components. 

\subsection{The proposed Transformer-based model: T4PdM}

The proposed model was developed based on the two datasets mentioned above. Independently, the MaFaulDa and the CWRU databases were randomly split into three disjoint sets: training, validation, and test. The split was set up to each set with the same fault proportion as the whole database. Therefore, 57.6\% of the data were separated for training, 18\% for validation, and 24.6\% for test.

The Transformer architecture is based on a multi-head attention mechanism that makes them particularly well suited to simultaneously represent each input sequence element by considering its context (future-past) using a positional-encoding layer, while multiple attention heads can consider different representation subspaces. This enables the recognition of multiple aspects of relevance between sequence input elements. In the case of time series, this may correspond to multiple patterns present in the signal for different time scales and periodicities \citep{Zerveas2020}. The original Transformer architecture is depicted in Figure 1. More details of the Transformer architecture can be found in \citet{Vaswani2017}.

\begin{figure}[h]\centering
	\includegraphics[scale=0.8]{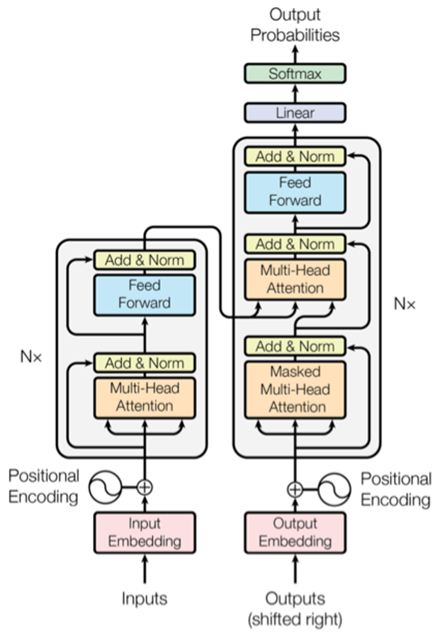}
	\caption{The original Transformer architecture \citet{Vaswani2017}.}\label{FIG_1}
\end{figure}

\begin{figure}[h]\centering
	\includegraphics[scale=0.8]{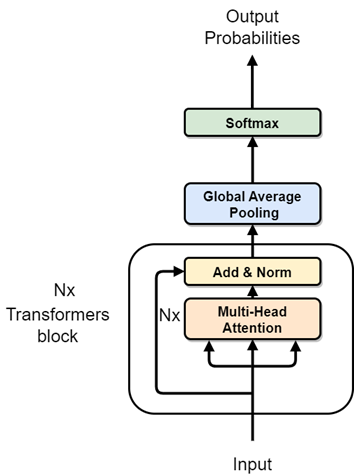}
	\caption{The proposed architecture of the T4PdM model developed in this work}\label{FIG_1}
\end{figure}

The Transformer layer is the main portion of the architecture of the proposed model, which is composed of layers of blocks stacked on top of each other, as shown in Figure 2. This paper presents a proposal change that makes it compatible with the vibration data in the domain frequency. For this end, the decoding and positional encoding (components of the original design) were not employed, as shown in Figure 2. The decoder module becomes unnecessary, since it is suitable for generative tasks, especially when dealing with problems involving sequence-to-sequence predictions, such as translation or summarization tasks in NLP. In addition, after the FFT, the vibration data that was in the time domain transforms into the frequency domain, being thus unnecessary the encoding of information about the relative or absolute position of the samples in the input. Therefore, the positional encoding was not added to the proposed model, thus not affecting the embeddings of the architecture.

The proposed model is composed of four Transformer encoder blocks and one-layer multi-head attention.  As a regularization parameter, a dropout layer with 0.5 units is applied followed by a global average pooling layer and a fully connected layer with Softmax activation function, as shown in Figure 2. This model becomes even more suitable when taking into account its simpler architecture in comparison with the original Transformer architecture, which makes it easier to train and deploy in production environment, which is essential in real world industry applications. T4PdM was implemented in Python using the Tensorflow framework with Keras.

\section{Results and discussion}

Several experiments were performed during the model development, using the test data for assessment. The first experiment applied a single layer of multi-head attention to one Transformer block, while the second experiment removed features of variance below 3.68e-5. The third experiment applied PCA to reduce dimensionality to 4,500 components. The fourth added three multi-head attention layers, and the fifth and last experiment increased from one to four the number of blocks of concatenated Transformers. Table 1 displays the evaluation performance of the five experiments described, where AUC ROC is Area Under Curve (AUC) of Receiver Operating Characteristics (ROC) and AUC PRC is AUC of Precision-Recall Curve (PRC). 

\begin{table}
\centering
\caption{Evaluation performance of five models applied for the MaFaulDa dataset}
\begin{tabular}{llllllll} 
\toprule
No. & Experiment Configuration & F1-Score & AUC ROC & AUC PRC & Recall  & Precision & Accuracy  \\ 
\hline
1   & Transformer              & 92.00\%  & 94.96\% & 87.43\% & 90.63\% & 96.18\%   & 98.80\%   \\
2   & Transformer + FS         & 99.71\%  & 99.70\% & 99.17\% & 99.45\% & 99.68\%   & 99.91\%   \\
3   & Transformer + PCA        & 99.71\%  & 99.74\% & 99.37\% & 99.52\% & 99.81\%   & 99.94\%   \\
4   & Transformer4m + PCA      & 99.86\%  & 99.86\% & 99.66\% & 99.74\% & 99.91\%   & 99.98\%   \\
5   & Transformer4b + PCA      & 99.86\%  & 99.92\% & 99.71\% & 99.85\% & 99.85\%   & 99.98\%   \\
\hline
\end{tabular}
\end{table}

The feature selection revealed a significant impact in the performance of the model, especially on the AUC PRC, which increased 13.42\%. PCA had a slightly contribution in the increasing of the performance. The best performance was achieved in the fifth experiment, which used PCA with four blocks of concatenated Transformers, as shown in Table 1. Therefore, the final architecture of the T4PdM model is the Transformer4b + PCA. The convergence of the model of the fifth experiment for the training and validation procedures obtained a steady state in 72 epochs.

The last layer of the proposed model was replaced to enable the training for the CWRU dataset.

\begin{figure}[!htb]\centering
	\includegraphics[scale=0.9]{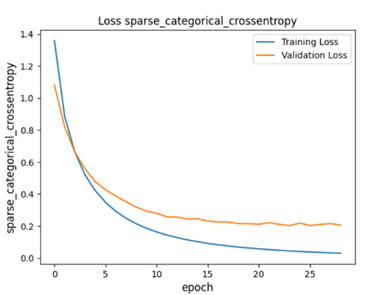}
	\caption{The curve of the loss during the training and validation applied to Transformer}\label{FIG_1}
\end{figure}
\begin{figure}[!htb]\centering
	\includegraphics[scale=0.9]{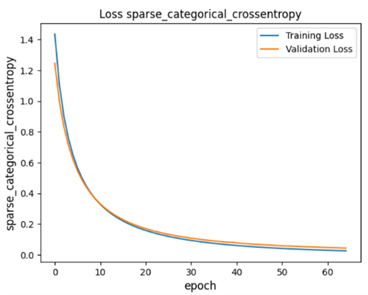}
	\caption{The curve of the loss during the training and validation applied to Transformer+FS}\label{FIG_1}
\end{figure}

\begin{figure}[!htb]\centering
	\includegraphics[scale=0.9]{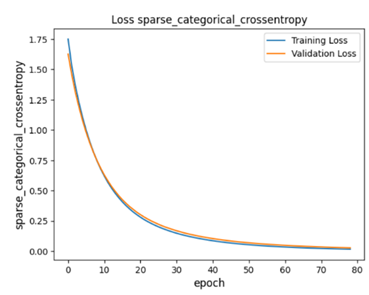}
	\caption{The curve of the loss during the training and validation applied to   Transformer+PCA}\label{FIG_1}
\end{figure}
\begin{figure}[!htb]\centering
	\includegraphics[scale=0.9]{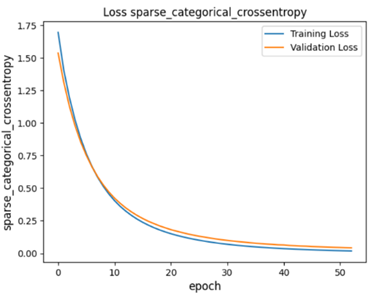}
	\caption{The curve of the loss during the training and validation applied to Transformer4m+PCA}\label{FIG_1}
\end{figure}
\begin{figure}[!htb]\centering
	\includegraphics[scale=0.9]{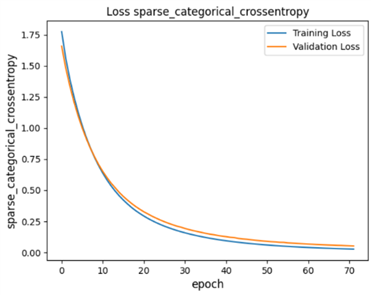}
	\caption{The curve of the loss during the training and validation applied to  Transformer4b+PCA}\label{FIG_1}
\end{figure}

Figure 3-7 shows the evolution of the losses of the training and validation processes for each model. All of them converged for very similar values of loss, which demonstrates that the models do not suffer from overfitting or underfitting. Despite the number of epochs being set to 100, the steady state was reached in 29 epochs for (3), 65 epochs for (4), 79 epochs for (5), 53 epochs for (6), and 72 epochs for (7).

\subsection{Evaluation Comparison}

After training and evaluating the proposed model, based on the configuration "Transformer4b + PCA" from Table 1, this work also analyzed its ability to learn from a new dataset. Therefore, the CWRU database was also used to train the model and perform a new test.

The model was able to accurately classify all classes in the MaFaulDa database with high accuracy even for the normal state, which presents the smallest number of samples. However, the model revealed a slight difficulty in the CWRU database. As depicted by Figure 9, 12.5\% (one sample) of the inner race samples were misclassified as ball fault, while the normal and outer race showed 100\% accuracy, despite the normal class still having fewer samples than the other classes. Overall, the model presented a better performance for the MaFaulDa than CWRU, as shown in Figure 9 in the confusion matrix. The main reason may be due to CWRU having much less data than MaFaulDa.

\begin{figure}[!htb]\centering
	\includegraphics[scale=0.8]{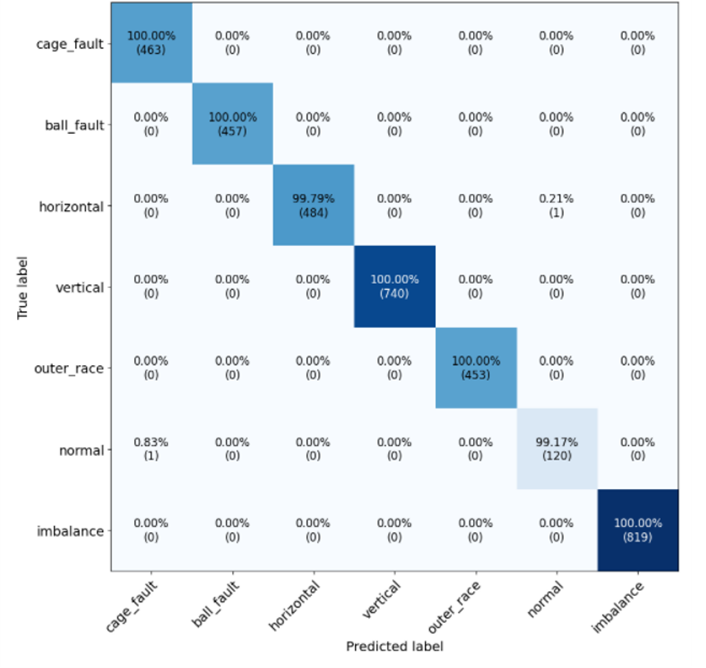}
	\caption{The proposed architecture of the T4PdM model developed in this work}\label{FIG_1}
\end{figure}

\begin{figure}[!htb]\centering
	\includegraphics[scale=0.8]{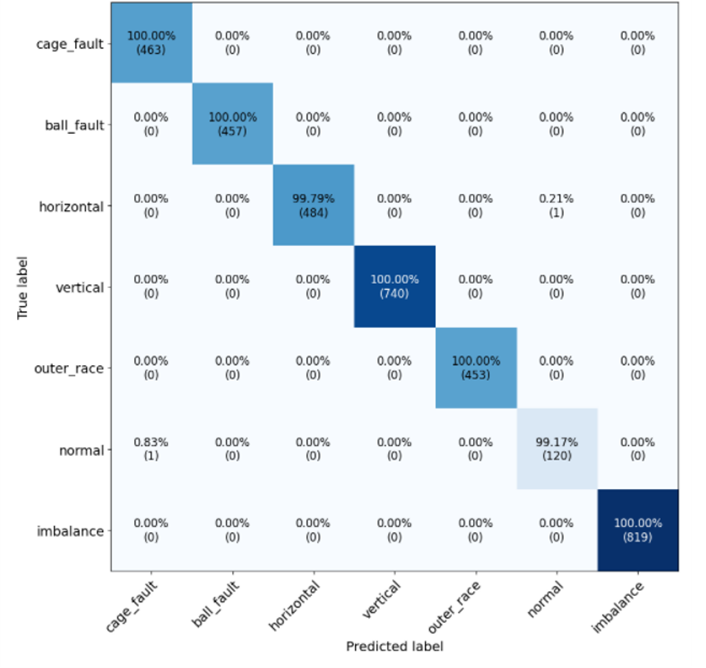}
	\caption{The proposed architecture of the T4PdM model developed in this work}\label{FIG_1}
\end{figure}

In the first experiment, the model revealed 100\% accuracy for classifying imbalance, vertical misalignment, outer race, and ball fault and 99\% accuracy for no fault, cage faults, horizontal misalignment, and ball fault for the MaFaulDa database. In the second experiment, the model revealed 100\% accuracy for classifying normal state and outer race and 99\% accuracy for ball fault and inner race for the CWRU database, as shown in Table 2 and 3.

\begin{table}
\centering
\caption{Evaluation performance of experiments for MaFaulDa}
\begin{tabular}{lllll} 
\toprule
Classes                 & F1       & Precision & Recall   & Accuracy  \\ 
\midrule
Cage fault              & 100.00\% & 99.78\%   & 100.00\% & 99.97\%   \\
Ball fault              & 100.00\% & 100.00\%  & 100.00\% & 100.00\%  \\
Horizontal mis. & 100.00\% & 100.00\%  & 99.79\%  & 99.97\%   \\
Vertical mis.   & 100.00\% & 100.00\%  & 100.00\% & 100.00\%  \\
Outer race              & 100.00\% & 100.00\%  & 100.00\% & 100.00\%  \\
Normal                  & 99.00\%  & 99.17\%   & 99.17\%  & 99.94\%   \\
Imbalance               & 100.00\% & 100.00\%  & 100.00\% & 100.00\%  \\
Average                 & 99.86\%  & 99.85\%   & 99.85\%  & 99.98\%   \\
\hline
\end{tabular}
\end{table}

\begin{table}[!htb]
\centering
\caption{Evaluation performance of experiments for CWRU}
\begin{tabular}{lllll} 
\toprule
Classes                 & F1       & Precision & Recall   & Accuracy  \\ 
\midrule
Ball\_fault & 93.00\%  & 87.50\%   & 100.00\% & 96.00\%   \\
Inner race  & 93.00\%  & 100.00\%  & 87.50\%  & 96.00\%   \\
Normal      & 100.00\% & 100.00\%  & 100.00\% & 100.00\%  \\
Outer\_race & 100.00\% & 100.00\%  & 100.00\% & 100.00\%  \\
Average     & 96.50\%  & 96.88\%   & 96.88\%  & 98.00\%   \\
\hline
\end{tabular}
\end{table}

As shown in Table 3, the model presented an overall accuracy of 98\% for the CWRU database, which highlights its capacity to be retrained for new data. It is also slightly superior to the work of \citet{Souza2021}, who has obtained 97.3\% accuracy using the CWRU database.

The proposed model was compared with other machine and deep learning models that have been developed in previous works applied to the MaFaulDa dataset. According to Table 4, the proposed model revealed the highest performance. 

\begin{table}
\centering
\caption{Evaluation performance of experiments for MaFaulDa}
\begin{tabular}{llllllll} 
\toprule
Reference               & Model    & Signal processing methods        & \begin{tabular}[c]{@{}l@{}}\textit{Average}\\ \textit{Accuracy}(\%)\textbf{\textit{}} \end{tabular} & Precision & \textit{F1- score} (\%)\textbf{\textit{}} & \begin{tabular}[c]{@{}l@{}}Recall \\\textit{\textbf{~}}\end{tabular}  \\ 
\hline
{}\citep{Ribeiro}                  & SBM + RF & Discrete Fourier Transform (DFT) & 96.43                                                                                               & -         & 99.24                                     & -                                                                     \\
{}\citep{Nath2021}                  & GRU      & DFC + Soft-DTW                   & 98.30                                         & -         & -                                         & -                                                                     \\
{}\citep{Marins2017}                  & SBM + RF & Discrete Fourier Transform (DFT) & 98.50                                                                                               & -         & -                                         & -                                                                     \\
{}\citep{Silva}                  & XGBoost  & Haar Wavelet                     & 99.16                                                                                               & -  & -                                         & -                                                                     \\
{}\citep{Souza2021}                  & PdM-CNN  & FFT                              & 99.58                                                                                               & 99.62     & 99.69                                     & 99.76                                                                 \\
\textit{Proposed model} & T4PdM    & FFT + PCA & 99.98                                                                                               & 99.85     & 99.86                                     & 99.85\\
\hline
\end{tabular}
\end{table}

The work of \citet{Silva} has used the XGBoost classifier and the Haar-wavelet transform, which obtained an accuracy of 99.16\%. The author has used the MaFaulDa database to model the experiment, and the vertical misalignment fault had the lowest accuracy of all failures. Authors in \citet{Ribeiro} have used the Similarity-Based Modelling (SBM) and Random Forest model, which obtained an accuracy of 96.43\%. The raw data was transformed into spatial and statistical properties (DFT, kurtosis, mean, and standard deviation) of each signal during the pre-processing step. While the work presented by \citet{Marins2017} has reported an accuracy of 98.50\%, similar to the previous work, it used similarity-based modeling (SBM) and the Random Forest model.

\section{Conclusion and future works}
\label{sec:others}

This work has proposed an intelligent diagnosis method based on Transformer Neural Network, called T4PdM. Through case studies of rotating machinery vibration datasets, it shows that the proposed method adaptively learns complex relationships from acquired data for various fault categories and is superior to state-of-the-art existing methods. From the results obtained, this method has demonstrated great generalization and excellent fault diagnosis abilities. The performance of the algorithm was superior to other approaches and achieved 99.98\% accuracy, which is a relatively high level of accuracy considering fault categorization with only one set of vibration sensors. The high capability of the T4PdM method was demonstrated using even less data and resources, and we have compared it to similar studies that have used the same databases, which could leverage industries to improve the performance and reliability of their manufactory processes by decreasing the number of unnecessary interventions and work accidents. The model was also applied to the CWRU database, which yielded an accuracy of 98\% in average, thus revealing the adaptability and applicability of the proposed method topology to other similar cases. 

This method has revealed great potential for application in identifying and classifying rotating machinery problems in industrial circumstances with varying severity levels, even when using only one vibration sensor and a small number of data, the FFT technique on the vibration sensor data, and the application of PCA for dimensionality reduction. Thus, the employment and improvement of the Transformer architecture for fault classification in rotating machinery has been shown to be feasible, viable, and effective.

As future works, an interesting research topic could be the application of the model on other industrial applications for the development of real-time fault classification in other different types of machines. In addition, a comparison of other methods in the literature for dimensionality reduction and their influence in the result could be carried out. Introducing convolution operations into the Transformer can also be explored aiming at improving the model’s performance, so that these hybrid models can be applied in the field of fault diagnosis. Another potential application is the implementation to detect the remaining useful life of parts in rotating machines and associate the Transformer model to the maintenance schedule management.

\section{Acknowledgment}

We gratefully acknowledge the support of SENAI CIMATEC AI Reference Center and the SENAI CIMATEC/NVDIA AI Joint Lab for scientific and technical support, as well as the SENAI CIMATEC Supercomputing Center for Industrial Innovation for granting access to the necessary hardware and technical support.



\bibliographystyle{unsrtnat}
\bibliography{artigoT4}  






\end{document}